\documentclass[lettersize,journal]{IEEEtran}
\usepackage[colorlinks,
            linkcolor=blue,
            anchorcolor=blue,
            citecolor=blue]{hyperref}
\usepackage{bm,amssymb,amsmath, amsfonts,booktabs,cleveref,multirow,adjustbox}
\usepackage{algorithm,algorithmicx}
\usepackage{array}
\usepackage[caption=false,font=normalsize,labelfont=sf,textfont=sf]{subfig}
\usepackage{textcomp}
\usepackage[section]{placeins}
\usepackage{stfloats}
\usepackage{url}
\usepackage{tikz}
\usepackage{verbatim}
\usepackage{tabularx}
\usepackage{epstopdf}
\usetikzlibrary{trees}
\usepackage{algpseudocode}
\usepackage{graphicx}
\def\BibTeX{{\rm B\kern-.05em{\sc i\kern-.025em b}\kern-.08em
    T\kern-.1667em\lower.7ex\hbox{E}\kern-.125emX}}
\usepackage{balance}

\begin{document}

\title{Interpretable Clustering Ensemble}
\author{Hang Lv, Lianyu Hu, Mudi Jiang, Xinying Liu, and Zengyou He
\thanks{The authors are with the School of Software, Dalian University of Technology, Dalian, China (e-mail: zyhe@dlut.edu.cn).
\par Manuscript recieved XXXX XX, 2025; revised XXXX XX, 2025.}}

\markboth{ IEEE TRANSACTIONS ON KNOWLEDGE AND DATA ENGINEERING, VOL. XX, NO. XX, XX 2025}%
{Shell \MakeLowercase{\textit{et al.}}: A Sample Article Using IEEEtran.cls for IEEE Journals}

\maketitle

\begin{abstract}
Clustering ensemble has emerged as an important research topic in the field of machine learning. Although numerous methods have been proposed to improve clustering quality, most existing approaches overlook the need for interpretability in high-stakes applications. In domains such as medical diagnosis and financial risk assessment, algorithms must not only be accurate but also interpretable to ensure transparent and trustworthy decision-making. Therefore, to fill the gap of lack of interpretable algorithms in the field of clustering ensemble, we propose the first interpretable clustering ensemble algorithm in the literature. By treating base partitions as categorical variables, our method constructs a decision tree in the original feature space and use the statistical association test to guide the tree building process. Experimental results demonstrate that our algorithm achieves comparable performance to state-of-the-art (SOTA) clustering ensemble methods while maintaining an additional feature of interpretability. To the best of our knowledge, this is the first interpretable algorithm specifically designed for clustering ensemble, offering a new perspective for future research in interpretable clustering.
\end{abstract}
\begin{IEEEkeywords}
Clustering ensemble, interpretable clustering, decision tree, unsupervised learning.
\end{IEEEkeywords}

\section{Introduction}

\IEEEPARstart{C}{lustering} analysis \cite{oyewole2023data} is an unsupervised learning issue in the field of data mining, which aims to partition data into different clusters by exploring its intrinsic structure. It has been widely applied in various domains, such as image segmentation \cite{Mittal2022}, social network analysis \cite{Socialnet}, and gene expression data analysis \cite{gene}. During the past decades, numerous clustering algorithms have been developed, including classic algorithms like \( k \)-means, DBSCAN, hierarchical clustering, and spectral clustering. However, each clustering algorithm has its own limitations and cannot always have the best performance across different data sets. For instance, the stability of \( k \)-means is greatly influenced by the pre-specified number of clusters \( k \) and the initial selection of centroids. These limitations restrict the generalization performance of individual clustering methods and make them less robust to noise and outliers.  

To improve both the clustering accuracy and stability, clustering ensemble techniques have emerged as a promising solution \cite{GOLALIPOUR2021104388}. By integrating multiple clustering results, the clustering ensemble method constructs a consensus partition that is expected to be more accurate and robust than its base learners. Existing studies have validated that ensemble strategies effectively mitigate issues related to parameter sensitivity and model assumption bias.  

However, existing approaches predominantly focus on optimizing the consensus partition to enhance clustering accuracy while neglecting the interpretability of ensemble results. The lack of interpretability severely constrains their deployment in high-stakes applications, where transparency is essential. However, the development of an interpretable clustering ensemble algorithm remains an open research issue in machine learning and data mining. 

In recent years, the development of interpretable clustering algorithms has emerged as a crucial research frontier in the field of machine learning \cite{Hu2024InterpretableCA}. Current interpretable clustering algorithms mainly focus explaining the clustering results of a specific algorithm (e.g., $k$-means) \cite{SHA} or directly generating clustering results in the form of an interpretable model (e.g., decision tree) \cite{bertsimas2021interpretable}. However, how to generate an interpretable clustering ensemble result remains unaddressed so far. 

Based on above observations, this paper proposes an interpretable clustering ensemble algorithm for the first time, which is named as ICE (Interpretable Clustering Ensemble). Essentially, the ICE algorithm treats each base partition as a categorical variable and seeks to construct a decision tree in the original feature space to make split decision explainable and meaningful. More precisely, the splitting criterion at each internal node is a predicate in terms of ``feature $\leq$ threshold'', which partitions the current sample set into two subsets. Therefore, we may consider each candidate splitting condition as a binary variable, whose goodness can be evaluated based on its statistical associations with base partition variables. Hence, during the top-down iterative tree construction process, we consider all possible candidate splitting conditions in the original feature space and the one with largest sum of chi-squared statistics is chosen. Each chi-squared statistic is calculated for assessing the association between the candidate splitting condition and each base partition. The tree growth process is terminated when we have obtained \( k \) leaf nodes, where \( k \) is a number of user-specified clusters. 

The experimental results demonstrate that compared to existing interpretable clustering methods, our algorithm improves the clustering accuracy while maintaining interpretability. Furthermore, compared to state-of-the-art (SOTA) clustering ensemble methods, our algorithm can achieve the same level clustering accuracy with an additional advantage in the sense that the generation of final partition is explainable.

In short, the main contributions of this paper are summarized as follows:
\begin{itemize}

    \item To our knowledge, this is the first piece of work on interpretable clustering ensemble in the literature.

    \item The proposed ICE algorithm is capable of generating a decision tree in the original feature space to illustrate how consensus clusters are produced. 

    \item Experimental results on multiple datasets demonstrate the competitive performance of our method in both clustering quality and interpretability.
\end{itemize}

The rest of this paper is organized as follows. Section 2 reviews related work on clustering ensembles and interpretable clustering. Section 3 describes the proposed algorithm. Section 4 presents experimental results and Section 5 concludes this paper.

\section{Related work}

\subsection{Clustering ensemble methods}

Clustering ensemble typically consists of two main phases: generating a set of base clustering results and combining them via a consensus function to obtain a robust final result. 
To date, numerous clustering ensemble algorithms have been proposed \cite{GOLALIPOUR2021104388,2011survey,BOONGOEN2018}.
Notable early contributions include HGPA \cite{Strehl2002}, HBGF \cite{fern2004solving}, EAC \cite{fred2005combining}, which laid the foundation for co-association matrix and hypergraph-based consensus techniques.  Subsequent research has further diversified consensus mechanisms, with methods such as RW \cite{abdala2010ensemble}, PA \cite{wang2009clustering} introducing different strategies for combining base clusterings. More recently, methods such as SPACE \cite{zhou2023active}, AWKEC \cite{li2024adaptive} and ICCM \cite{khedairia2022multiple} further enhanced consensus strategies through active learning, kernal learning and iterative combining respectively.

Roughly, existing clustering ensemble algorithms can be broadly categorized into three types based on the consensus function employed.
The first category is co-association matrix based methods \cite{fred2005combining,wang2009clustering,li2024adaptive}, which build a similarity matrix that records the frequency of sample pairs being assigned to the same cluster across all base clusterings.  This matrix is then used as input for consensus partition generation, typically via similarity-based techniques such as hierarchical clustering, spectral methods, or graph-based strategies.
The second category is hypergraph partitioning-based methods \cite{Strehl2002,fern2004solving,abdala2010ensemble}, where each cluster in a base clustering is treated as a hyperedge connecting multiple samples, and the ensemble task is transformed into a hypergraph partitioning problem to derive the consensus partition.
The third category is relabeling-based methods \cite{khedairia2022multiple,ayad2010voting,saeed2012voting}, which aim to align the labels of different base clusterings and then aggregate them using voting or optimization strategies.  These methods typically rely on solving the label correspondence problem among base partitions.

Although many clustering ensemble methods have been proposed, most of them emphasize on improving the clustering accuracy, with little consideration of interpretability. In this paper, we propose an interpretable clustering ensemble method to bridge this gap.

\subsection{Interpretable clustering}

In recent years, clustering has been widely adopted in domains such as healthcare, finance, and social sciences. However, its black-box nature often hinders deployment in settings where interpretability is essential, thereby motivating increasing research interests on interpretable clustering methods. A variety of interpretable clustering algorithms have been proposed \cite{Hu2024InterpretableCA}, employing different models, including rules \cite{carrizosa2023clustering}, decision trees \cite{SHA,fraiman2013interpretable,hwang2023xclusters}, prototypes \cite{carrizosa2022interpreting}, convex-polyhedral \cite{lawless2023cluster} and description \cite{davidson2018cluster}.   Among these, decision tree based algorithms are particularly popular due to their intuitive interpretability, as each cluster assignment can be explained by a path from the root to a leaf node.   In this work, we adopt decision tree as the interpretable model to provide transparent explanations of clustering results.

In addition, interpretable clustering methods can be categorized based on the stage of the clustering process at which interpretability is introduced: pre-clustering, in-clustering, and post-clustering. Pre-clustering methods \cite{bonifati2022time2feat,svirsky2023interpretable,effenberger2021interpretable} enhance interpretability by performing feature extraction or selection before clustering, generating more meaningful inputs for the model. In-clustering methods \cite{fraiman2013interpretable,hwang2023xclusters,saisubramanian2020balancing} incorporate interpretability directly into the clustering algorithm by jointly optimizing for clustering quality and interpretability, resulting in inherently explainable outcomes. Post-clustering methods \cite{SHA,carrizosa2023clustering,carrizosa2022interpreting} generate explanations after clustering is completed, using interpretable models to approximate the results of black-box algorithms and reveal the underlying decision logic.

From the perspective of the above categorization, our method belongs to the post-clustering category, as it explains given clustering results by constructing a binary decision tree. Existing post-clustering methods such as IMM \cite{pmlr-v119-moshkovitz20a} focus on explaining an individual base partition, whereas how to integrate multiple base partitions into an interpretable consensus result remains an unexplored problem.

\subsection{Significance testing for clustering}

Statistical testing has emerged as a key approach in clustering analysis to assess whether identified structures are statistically significant. Related significance testing problems can be grouped into four categories: clusterability test \cite{adolfsson2019cluster}, which evaluates whether a data set can be divided into multiple meaningful groups of samples; cluster number test \cite{maitra2012bootstrapping}, which determines the appropriate number of clusters in the data set; partition test \cite{helgeson2021nonparametric}, which computes $p$-value to assess the statistical significance of a given partition (i.e., a specific set of clusters); and individual cluster test \cite{xie2021significant}, which evaluates the statistical significance of a specific cluster.

Recently, significance testing has also been applied to interpretable clustering \cite{hu2025interpretable,hu2025significance,he2025significance}. 
These methods employ statistical significance tests to compute $p$-value that guide the growth and stopping criteria of decision trees. This ensures that each node in the decision tree is statistically reliable, thereby enhancing the interpretability.

Note that significance testing methods have been applied to the interpretable cluster analysis of categorical data and sequential data in  these existing research efforts. However, the use of such testing-based methods for generating consensus partition in clustering ensemble setting is still not investigated. 

\section{Methods}

\subsection{Preliminaries}\label{3.1}

\begin{figure*}[!t]
\centering
\includegraphics[width=\textwidth]{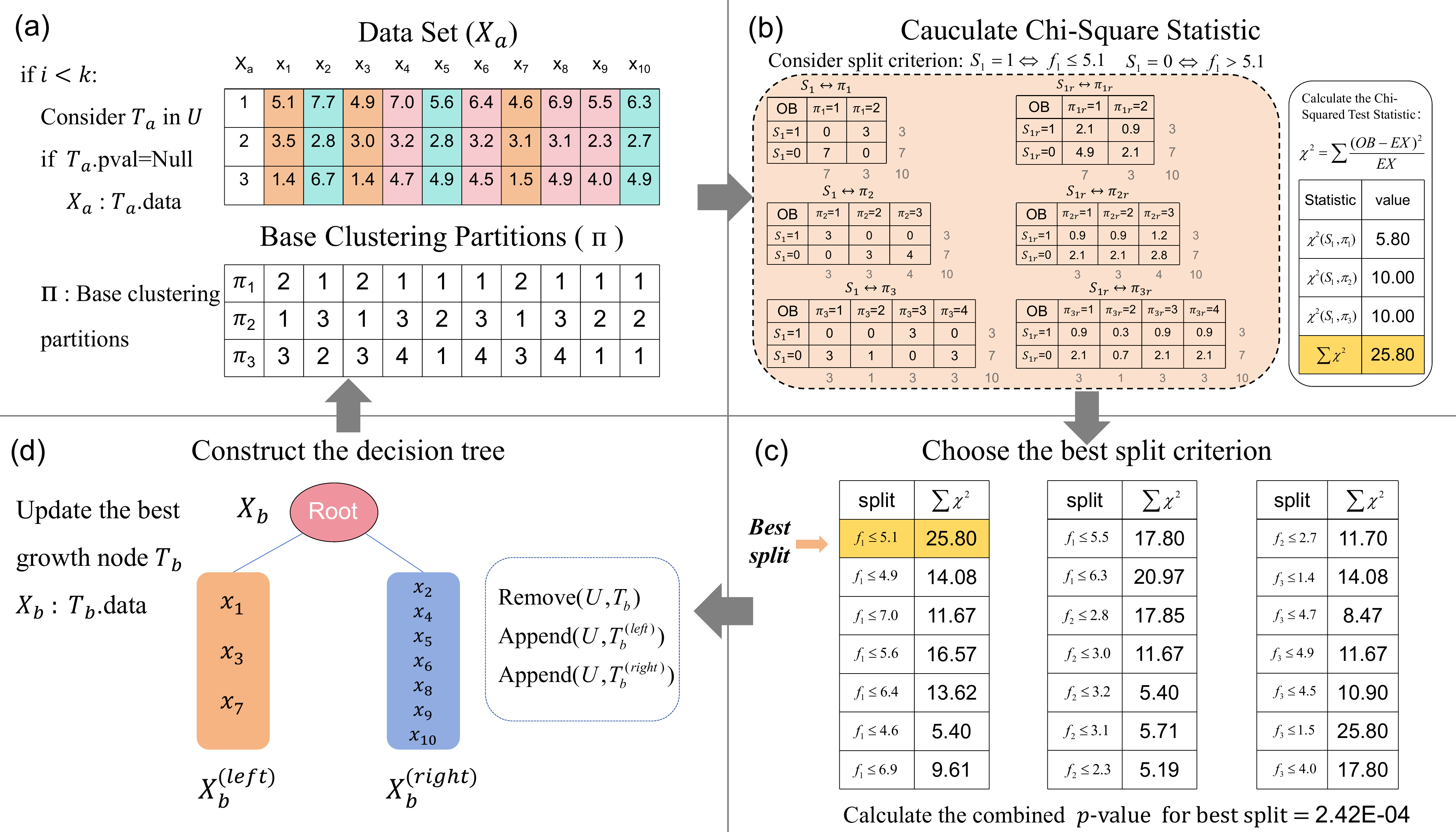}
\caption{ Illustration of the proposed ICE method: (a) Initially, the algorithm processes nodes $T_a$  as demonstrated through a toy dataset containing 10 samples and 3 features.  (b) For each potential split, such as  $f_{1} \leq 5.1$, the data is divided into two subsets, and the chi-squared statistic between this binary partition and each base partition is computed and aggregated.  (c) The split achieving the smallest $p$-value is identified as the best split for the node. (d) Subsequently, the leaf node with the minimal $p$-value is selected for growth, after which $U$ is updated. The above process is repeated for $k-1$ times.}
\label{fig:a} 
\end{figure*}

Consider an original data set $X = \{x_1,x_2, \ldots, x_N\} \subseteq \mathbb{R}^M$ is composed of $N$ samples with $M$ features $ \{ f_1,f_2, \ldots, f_M \}$, where each sample $x_i \in \mathbb{R}^M$ is an $M$-dimensional feature vector. The $j$-th feature value of the $i$-th sample is denoted by $f_{j}(x_{i})$. For any subset $X^{'} \subseteq X$ and the $j$-th feature ($1 \leq j \leq M$), let $\mathcal{V}_j(X^{'})$ denote the set of distinct feature values that appear in  $X'$. Let \(\Pi = \{\pi_1, \pi_2, \ldots, \pi_{c}\}\) represent the set of \(c\) base clustering partitions, where in each base partition \(\pi_t\) \((1 \leq t \leq c)\) every sample \(x_i \in X\) is associated with a cluster label. Suppose the base partition $\pi_t$ is composed of $p_t$ clusters, we can regard this partition as a categorical variable with $p_t$ categories. 

When constructing the clustering decision tree, the root node, initialized as $T$, contains the entire dataset. We use a set $U$ to store candidate nodes that can be further divided. The clustering decision tree partitions the original data set $X$ into $k$ clusters, where $k$ is a user-specified parameter and each cluster is determined by a leaf node.
Given a node $T_a$, let \( X_a \subseteq X \) denote the subset of samples in $T_a$.  To divide $T_a$, a candidate split takes the following form: $f_j \leq v_j$, where $f_j$ can be arbitrarily selected from all \( M \) features, and the threshold \( v_j \) is freely chosen from \( \mathcal{V}_j(X_a) \). We can define a binary variable $S_{j}$ for the above candidate split:  $S_{j}=1 $ if $ f_{j} \leq v_{j}$ is true and  $S_{j}=0$ otherwise. 
Based on this candidate split, \( X_a \) is partitioned into two subsets:
\begin{align}
X_a^{(\text{left})} &= \left\{ x_i \in X_a \mid f_j(x_i) \leq v_j \right\}, \label{eq:left} \\
X_a^{(\text{right})} &= \left\{ x_i \in X_a \mid f_j(x_i) > v_j \right\}. \label{eq:right}
\end{align}
\subsection{Overview of ICE}

\indent
The ICE algorithm is designed to construct a decision tree that generates meaningful and interpretable clustering results by integrating multiple base partitions. As shown in Fig.~\ref{fig:a}, the ICE algorithm initiates with the root node containing the complete dataset, which is simultaneously added to the candidate splitting node set $U$. The sequential steps are as follows: (a) First, it checks whether the number of leaf nodes in the current tree is less than the predefined threshold;    if so,  nodes in $U$ whose best split has not been calculated are processed by utilizing their associated datasets and base clustering information.    (b) Subsequently, for each candidate splitting node, potential splits are evaluated by partitioning the data into two subsets, where chi-squared statistics between each binary split and all base partitions are calculated and summed to quantify split quality.    (c) The split with the smallest $p$-value is determined for each candidate node.     (d) Finally, the node exhibiting the smallest $p$-value is divided into two leaf nodes, followed by updating $U$, after which the algorithm  returns to step (a) until termination criteria are met.

\subsection{The ICE algorithm}

\subsubsection{Split criteria}\label{3.4}

Each split divides samples in current node into two subsets, which can be regarded as a binary variable. We treat each base clustering outcome as a categorical variable and assesses statistical associations through chi-squared test between base partitions and the candidate split. For the $c$ base partitions, the algorithm independently computes chi-squared statistics between each base partition and the split, then sum these values to obtain an aggregated statistic.    Higher  values indicate stronger correlations between the candidate split and multiple base clustering results, thereby indicating a better split. 

To be specific, when performing chi-squared test between a candidate split $S_j$ and a base partition $\pi_t$, we first construct a contingency table as shown in Table~\ref{table1}.

\begin{table}[!t]
\caption{The contingency table for variables $S_j$ and $\pi_t$.}
\renewcommand{\arraystretch}{1.5}
\label{table1}
\centering
\small  
\begin{tabularx}{0.48\textwidth}{c *{4}{>{\centering\arraybackslash}X}} 
\toprule
& $\pi_t=1$ & \ldots & $\pi_t=p_t$ & Total \\
\midrule 
$S_j=1$ & $N_{11}^{(j,t)}$ & $N_{1l}^{(j,t)}$ & $N_{1p_t}^{(j,t)}$ & $N_{1*}^{(j,t)}$ \\
$S_j=0$ & $N_{01}^{(j,t)}$ & $N_{0l}^{(j,t)}$ & $N_{0p_t}^{(j,t)}$ & $N_{0*}^{(j,t)}$ \\
Total   & $N_{*1}^{(j,t)}$ & $N_{*l}^{(j,t)}$ & $N_{*p_t}^{(j,t)}$ & $N^{(j,t)}$ \\
\bottomrule
\end{tabularx}
\end{table}
Based on the contingency table, we can first calculate the expected frequency as:

\begin{equation}
E_{sl}^{(j,t)}=\frac{N_{s*}^{(j,t)}\cdot N_{*l}^{(j,t)}}{N^{(j,t)}}.
\end{equation}

For each pair ($S_j,\pi_t$), the chi-squared statistic can be computed as:

\begin{equation}
\chi^2(S_j,\pi_t)=\sum \limits_{s=0}^1\sum \limits_{l=1}^{p}\frac{(N_{sl}^{(j,t)}-E_{sl}^{(j,t)})^2}{E_{sl}^{(j,t)}}. \label{eq2}
\end{equation}

Then, we calculate the sum of the chi-squared statistics, which is used to assess the goodness of the candidate split as follows:
\begin{equation}
\chi^2(S_j)=\sum \limits_{t=1}^{c}\chi^2(S_j,\pi_t). \label{eq3}
\end{equation}

However, the sum of statistics is only able to choose the best split for one fixed node. When comparing splits from multiple candidate splitting nodes, it may be not appropriate since the number of remaining cluster labels can be different across different nodes. Therefore, we further calculate the $p$-value based on the sum of chi-squared statistics under the independence assumption \cite{agresti2013categorical}. That is,  if we assume that $c$ chi-squared variables are independent, then the sum of statistics in Equation \eqref{eq3} follows a chi-squared distribution as well \cite{agresti2013categorical}. After calculating the $p$-value, then can select the candidate splitting node with the minimal $p$-value to grow the tree. 

\begin{algorithm}[!t]
    \renewcommand{\algorithmicrequire}{\textbf{Input:}}
    \renewcommand{\algorithmicensure}{\textbf{Output:}}
    \caption{Interpretable Clustering Ensemble}
    \label{ICE}
    \begin{algorithmic}[1]
        \Require Input data set $X$, the ensemble size $c$, base clustering partitions $\Pi$, the maximum number of leaf nodes $k$ 
        \Ensure An unsupervised clustering decision tree $T$ 
    
        \State \( T \gets \text{node(data}=X \text{, statisitc}=\text{Null, } \text{pval}=\text{Null, feature}=\text{Null, value}=\text{Null, left}=\text{Null, right}=\text{Null}) \) \label{line1} \label{a1}
        \State Initialize list $U \gets [\,T]$ \label{line:start}, $i=1$
        
        \While{$i < k$} \label{a3}
            \For{$T_a$ in $U$}
                \If{$T_a$.pval=Null}
                \State $S_{j^*}\gets $OptimalSplit($T_a\text{.data},\Pi$)
                \State Update $T_a.\text{pval}, T_a.\text{feature}, T_a.\text{value}$ based on $S_{j^*}$
                \EndIf
            \EndFor
            \State Update $T_b$ which has the minimal $p$-value in $U$ \label{sel}
            \State Initialize the left and right subtrees based on the optimal split of $T_b$
            \State Update $U$, $i=i+1$
        \EndWhile \label{while:end}
        \State \Return $T$
        \Function{OptimalSplit}{$X_a, \Pi$}\label{os}
            \For{ $j \in [1, \ldots , M]$}
                \State $X_{sorted} \gets \text{Sort}(X_a, j, \text{ascending})$ \label{sort} 
                \State Initialize $c$ independent $2 \times p_t$ contingency tables
                \For{every $v_j \in \mathcal{V}_j(X_{sorted})$}
                \State Obtain subsets $X_a^{(\text{left})}$ and $X_a^{(\text{right})}$ based on $S_j: f_j \leq v_j$
                \If {$\lvert X^{(left)}_{a}\rvert<5 \textbf{ or } \lvert X_{a}^{(right)}\rvert<5$}\
                \State continue
                \EndIf
                \State  Update the contingency tables incrementally 
                \State Update $S_{j^*}$ with the $ \chi^2(S_{j})$
                \EndFor

            \EndFor
        \State \Return $S_{j^*}$
        \EndFunction \label{os1}
    \end{algorithmic}
\end{algorithm}
\subsubsection{The details of proposed algorithm}
The proposed ICE algorithm integrates multiple base partitions into a unified decision tree structure in the original feature space, ensuring both interpretability and accuracy. Based on the concepts and symbols defined in Section~\ref{3.1}, we formally present the algorithm in Algorithm~\ref{ICE}.

The tree construction process follows an iterative greedy strategy by dynamically selecting optimal splits for hierarchical data partitioning. The root node $T$ is initialized with the input data set $X$. The main iterative loop contains two main steps. 

The first step is to find the best split of the candidate node whose best split has not been calculated in the set $U$. For a candidate node $T_a$, the optimal split $S_{j^*}$ is identified by the OptimalSplit function (Lines~\ref{os}$\sim$\ref{os1}). The split criteria have been introduced in Section~\ref{3.4}. However, our method needs to consider all possible candidate splits, which can be computationally intensive. So we consider optimizing the process by proposing a strategy based on sorting and incremental updates.  For each feature, we first sort its feature values in an ascending order (Line~\ref{sort}) and then consider candidate splits in an increasing order of feature values.  For candidate splits with the same feature, we only need to modify the contingency table incrementally, avoiding the need to regenerate the contingency table each time.

The another step is to update the splitting node in $U$, i.e., node $T_b$ with the minimal $p$-value (Line~\ref{sel}). The best split of the expansion node is used to build subtrees. These child nodes inherit subsets of $T_b$.data and are appended to $U$ for future evaluation while $T_b$ is removed from $U$.

The stopping criterion is defined as the termination of growth when reaching the prespecified leaf node number $k$. Since each split generates two new leaf nodes from one parent node, the loop runs $k$-1 times. The final tree, rooted at $T$, is returned. Each internal node contains its splitting rule, while leaf nodes represent clusters of the data.

\subsubsection{Time complexity analysis}
The time complexity of Algorithm~\ref{ICE} is analyzed as follows. The main loop (Lines~\ref{a3}$\sim$\ref{while:end}) iterates until $k$ leaf nodes are created. 

When $i=1$, the OptimalSplit function (Lines~\ref{os}$\sim$\ref{os1}) evaluates all candidate splits for node $T$ with $N$ samples. First, sorting $M$ features requires  $O(M\cdot N\cdot lgN)$ time. Then, the construction of $c$ initial  contingency tables for one feature needs $O (c \cdot N)$ time. After that, we can incrementally update the contingency tables for subsequent candidate splits in $O (N \cdot R)$, where $ R=2\sum \limits_{t=1}^c p_t$. Considering all $M$ features, the discovery of best split takes $O(M \cdot N \cdot R)$ time.  Therefore, the time complexity of the OptimalSplit function is $O (M \cdot N \cdot (lgN+R))$. 

At the root node, we need to run OptimalSplit function one time. After that, during the iteration procedure, we just need to  run OptimalSplit function to obtain the best split for each of two newly generated leaf nodes. Therefore, after generating $k$ leaf nodes, the OptimalSplit function is executed $2k-3$ times. Therefore, the overall time complexity of the ICE algorithm is at most $O (k \cdot M \cdot N \cdot (lgN+R))$.

\section{Experiments}
\subsection{Baselines}

In this section, we compare ICE with the following clustering algorithms on 34 data sets.
The baseline clustering algorithms are as follows:
\begin{itemize}

\item[$\bullet$] \textbf{KM}. The $k$-means algorithm, which is used to produce the base partitions for clustering ensemble algorithms. 

\item[$\bullet$] \textbf{IMM} \cite{pmlr-v119-moshkovitz20a}. It is an explainable $k$-means clustering algorithm that builds a decision tree using an iterative mistake minimization strategy.

\item[$\bullet$] \textbf{SHA} \cite{SHA}. It aims to minimize the $k$-means cost while constructing a shallow decision tree by incorporating penalty terms into the loss function.

\item[$\bullet$] \textbf{LWEA and LWGP} \cite{huang2017locally}. They are both locally weighted clustering ensemble algorithms.   LWEA applies an agglomerative consensus clustering method based on the local weighting strategy, while LWGP employs the local weighting strategy to enhance the graph partitioning process. 

\item[$\bullet$] \textbf{ECCMS} \cite{jia2023ensemble}. It improves the co-association matrix for clustering ensemble using high-confidence information propagation. 

\item[$\bullet$] \textbf{ECPCS-HC and ECPCS-MC} \cite{huang2018enhanced}. These two clustering ensemble algorithms enhance the clustering quality by propagating cluster similarities.
ECPCS-HC employs a hierarchical consensus function for sample-level aggregation, while ECPCS-MC uses a meta-cluster-based consensus function for cluster-level partitioning . 

\end{itemize}

\subsection{Data Sets and Experiment Setup}
\begin{table}[!t]
\caption{The characteristics of the data sets used in the experiment. \#Instances denotes the number of the samples within each data set, where the number in parentheses indicates the total number of samples with missing values (if any). \#Features is the number of features in the data set while \#classes represents the number of distinct clusters in the data set. }
\renewcommand{\arraystretch}{1}
\label{table2}
\centering
\setlength{\tabcolsep}{8pt}
\begin{tabularx}{0.45\textwidth}{cccc}
\toprule
Data sets&\#Instances &\#Features &\#Classes \\
\midrule
Appendicitis&106&7&2 \\
Banana&5300&2&2 \\
Bands&365(539)&19&2\\
Banknote&1372&4&2\\
Cervical Cancer &72&19&2\\
Glass&214&9&6\\
Haberman&306&3&2\\
Ionosphere&351&34&2\\
Iris&150&4&3\\
Maternal &1014&6&3\\
Movement&360&90&15\\
Newthyroid&215&5&3\\
NHANES&2278&7&2\\
Page-blocks&5472&10&5\\
Parkinsons&195&22&2\\
Penbased&10992&16&10\\
Phoneme&5404&5&2\\
Pima&768&8&2\\
Ring&7400&20&2\\
Satimage&6435&36&7\\
Seeds&210&7&3\\
Segment&2310&19&7\\
Sonar&208&60&2\\
Spambase&4601&57&2\\
Spectfheart&267&44&2\\
Texture&5500&40&11\\
User knowledge &403&5&4\\
Vehicle&846&18&4\\
Wdbc&569&30&2\\
Wine&178&13&3\\
Winequality-red&1599&11&6\\
Wisconsin&683(699)&9&2\\
Yeast&1484&8&10\\
\bottomrule
\end{tabularx}
\end{table}

\indent

We compare different algorithms on 33 data sets, all of which are sourced from the UCI \cite{dua2017uci} repository and the KEEL-dataset repository \cite{derrac2015keel}. All features in these datasets are numeric, and samples with missing values are discarded during the data preprocessing phase. The details of the data sets are summarized in Table~\ref{table2}.

We first employ the $k$-means algorithm to generate base partitions using the standardized input data. The number of clusters for each base partition is randomly selected within the range [$k, 3k$], where $k$ denotes the true number of clusters of the input data set. In this study, we set the ensemble size $c=30$  to construct a set of base partitions $\Pi$ and execute 10 independent runs to obtain the average value of each performance indicator for comparing different clustering ensemble methods. Our ICE method is configured such that the maximum number of leaf nodes matches the actual number of clusters in the data set. All hyperparameters of the comparison methods are configured in accordance with the recommendations specified in their original papers. 

The clustering quality of each algorithm is evaluated using three metrics: Purity, F1-score, and Normalized Mutual Information (NMI). To compare the interpretability of different interpretable clustering algorithms, we use the maximum depth of the tree (maxDepth) and the average depth of the leaf nodes (avgDepth) as the performance metric. 

\begin{table*}[!t]
\caption{The comparison of different clustering algorithms in terms of Purity.}
\renewcommand{\arraystretch}{1}
\label{table3}
\centering
\begin{tabularx}{0.8\textwidth}{c|c|c|cc|ccccc}
\hline
Data set&ICE&KM&IMM&SHA&LWEA&LWGP&ECCMS&ECPCS-HC&ECPCS-MC \\\hline
Appendicitis&0.8491  & 0.8057  & 0.8283  & 0.8491  & 0.8236  & 0.8104  & 0.8236  & 0.8066  & 0.8019 \\ 

        Banana&0.5641  & 0.5664  & 0.5594  & 0.5604  & 0.5533  & 0.5581  & 0.5601  & 0.5568  & 0.5669  \\ 
         Bands & 0.6400  & 0.6301  & 0.6301  & 0.6301  & 0.6312  & 0.6367  & 0.6334  & 0.6411  & 0.6411   \\ 
         Banknote & 0.6184  & 0.5587  & 0.6356  & 0.6450  & 0.5822  & 0.5923  & 0.6117  & 0.5714  & 0.5679   \\ 
         Cervical Cancer & 0.7500  & 0.7861  & 0.7569  & 0.8056  & 0.7833  & 0.7486  & 0.7833  & 0.7917  & 0.7778   \\ 
         Glass  & 0.5070  & 0.5491  & 0.5444  & 0.5439  & 0.5472  & 0.5575  & 0.5495  & 0.5453  & 0.5458   \\ 
         Haberman & 0.7353  & 0.7353  & 0.7369  & 0.7353  & 0.7412  & 0.7373  & 0.7428  & 0.7490  & 0.7405   \\ 
         Ionosphere & 0.6980  & 0.7066  & 0.6923  & 0.6980  & 0.7197  & 0.7071  & 0.7185  & 0.7060  & 0.7094   \\ 
         Iris & 0.8640  & 0.8287  & 0.7773  & 0.8200  & 0.8327  & 0.8307  & 0.8260  & 0.8267  & 0.8247   \\ 
         Maternal  & 0.5849  & 0.5410  & 0.5623  & 0.5286  & 0.5626  & 0.5648  & 0.5625  & 0.5250  & 0.5752   \\ 
         Movement & 0.4169  & 0.4722  & 0.4083  & 0.4514  & 0.4969  & 0.5036  & 0.4975  & 0.4981  & 0.4969   \\ 
         Newthyroid & 0.8744  & 0.8744  & 0.8744  & 0.8744  & 0.8800  & 0.8828  & 0.8809  & 0.8781  & 0.9005   \\ 
         NHANES & 0.8402  & 0.8402  & 0.8402  & 0.8402  & 0.8402  & 0.8402  & 0.8402  & 0.8402  & 0.8402   \\ 
         Page-blocks & 0.9188  & 0.9127  & 0.9037  & 0.9086  & 0.9207  & 0.9126  & 0.9205  & 0.9147  & 0.9109   \\ 
         Parkinsons & 0.7538  & 0.7538  & 0.7538  & 0.7538  & 0.7538  & 0.7538  & 0.7538  & 0.7538  & 0.7538   \\ 
         Penbased & 0.6457  & 0.7117  & 0.6960  & 0.6478  & 0.7783  & 0.8090  & 0.7648  & 0.7562  & 0.7723   \\ 
         Phoneme & 0.7065  & 0.7065  & 0.7218  & 0.7065  & 0.7065  & 0.7065  & 0.7065  & 0.7067  & 0.7065   \\ 
         Pima & 0.6521  & 0.6738  & 0.6660  & 0.6510  & 0.6513  & 0.6690  & 0.6533  & 0.6639  & 0.6646   \\ 
         Ring & 0.6053  & 0.7568  & 0.6104  & 0.6117  & 0.7606  & 0.7895  & 0.7638  & 0.7835  & 0.7791   \\ 
         Satimage & 0.7413  & 0.7413  & 0.7169  & 0.7142  & 0.7620  & 0.7073  & 0.6251  & 0.7624  & 0.7194   \\ 
         Seeds & 0.8390  & 0.9190  & 0.8843  & 0.8667  & 0.8819  & 0.8762  & 0.8643  & 0.8719  & 0.8486   \\ 
         Segment & 0.5413  & 0.5834  & 0.6152  & 0.5810  & 0.5528  & 0.5726  & 0.5353  & 0.5552  & 0.5552   \\ 
         Sonar & 0.5514  & 0.5341  & 0.5486  & 0.5663  & 0.5409  & 0.5361  & 0.5438  & 0.5361  & 0.5341   \\ 
         Spambase & 0.6060  & 0.6060  & 0.6507  & 0.6060  & 0.6060  & 0.6060  & 0.6060  & 0.6060  & 0.6060   \\ 
         Spectfheart & 0.7940  & 0.7940  & 0.7940  & 0.7940  & 0.7940  & 0.7940  & 0.7940  & 0.7940  & 0.7940   \\ 
         Texture & 0.6159  & 0.6314  & 0.6082  & 0.6010  & 0.6899  & 0.6582  & 0.6991  & 0.6587  & 0.6524   \\ 
         User knowledge & 0.5730  & 0.4553  & 0.5146  & 0.5050  & 0.5065  & 0.5241  & 0.5184  & 0.4888  & 0.4928   \\
         Vehicle & 0.3677  & 0.3674  & 0.3877  & 0.3830  & 0.3694  & 0.3937  & 0.3729  & 0.3770  & 0.3583   \\ 
         Wdbc & 0.8757  & 0.9088  & 0.8914  & 0.8998  & 0.8975  & 0.9151  & 0.8905  & 0.9060  & 0.9042   \\ 
         Wine & 0.9101  & 0.9652  & 0.9084  & 0.9101  & 0.9612  & 0.9685  & 0.9590  & 0.9685  & 0.9680   \\ 
         Winequality-red & 0.5433  & 0.5495  & 0.5321  & 0.5265  & 0.5463  & 0.5387  & 0.5430  & 0.5260  & 0.5366   \\ 
         Wisconsin & 0.9297  & 0.9564  & 0.9297  & 0.9297  & 0.9706  & 0.9704  & 0.9707  & 0.9703  & 0.9685   \\ 
         Yeast & 0.5255  & 0.5461  & 0.5313  & 0.5279  & 0.5607  & 0.5503  & 0.5606  & 0.5473  & 0.5453   \\ \hline
         Avg &  0.6860  & 0.6960  & 0.6882  & 0.6871  & 0.7032  & 0.7037  & 0.6993  & 0.6995  & 0.6988  \\ 
\hline
\end{tabularx}
\end{table*}
\subsection{Clustering quality comparison}

\begin{figure}[!t]
\centering
\includegraphics[height=0.6\textheight,width=\columnwidth]{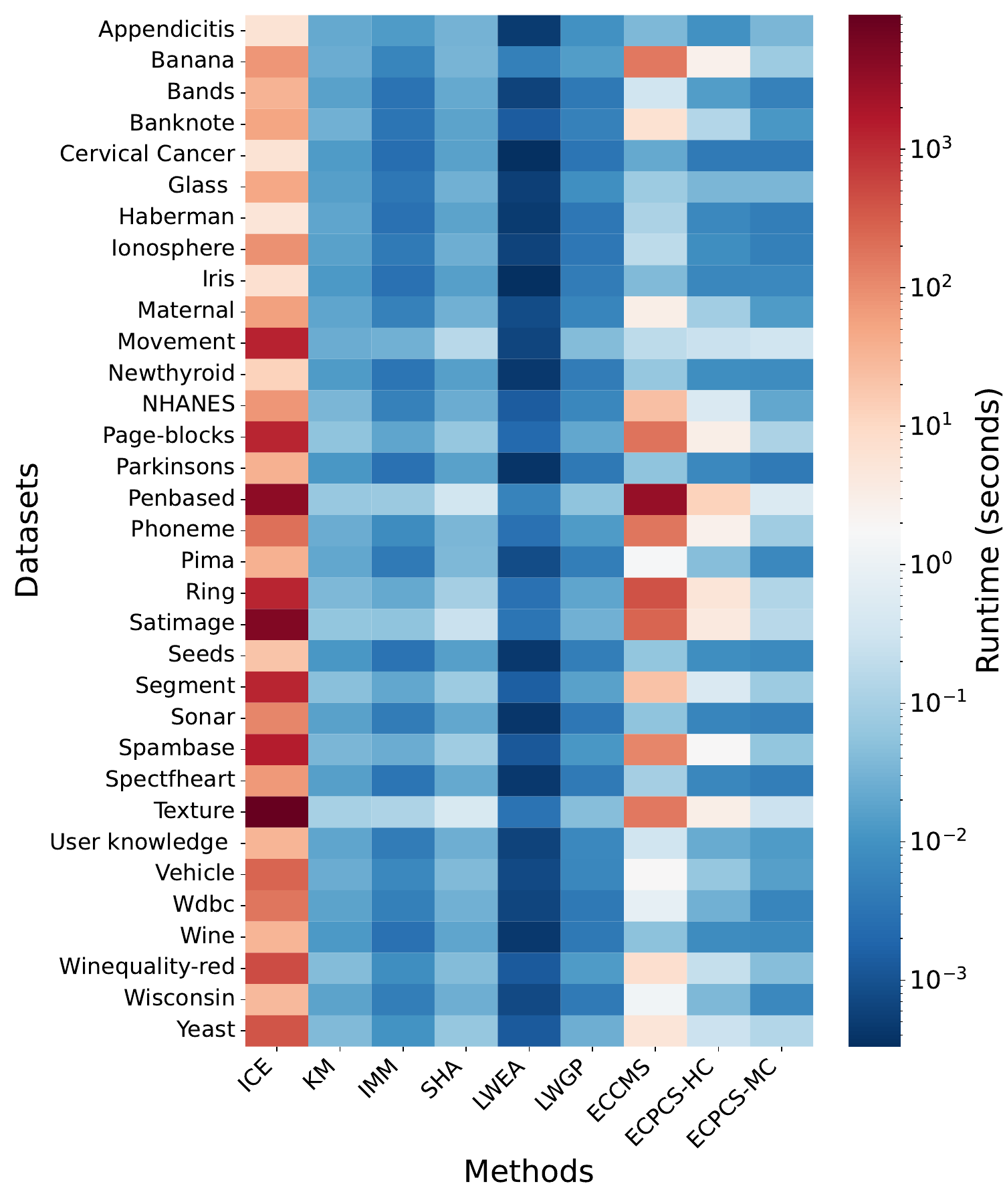}
\vspace{-10pt}
\caption{Comparison of running times (in seconds) of all algorithms on 33 datasets. All experiments are conducted on a PC with Intel(R) Core(TM) i5-11400F CPU 2.60GHz and 16G Memory.}
\label{figb}
\end{figure}

The experimental results in terms of Purity, F1-score and NMI are shown in Table~\ref{table3},~\ref{table4} and~\ref{table5}.
The execution time, recoreded in seconds, are illustrated in Fig.~\ref{figb}.   
Some important observations are summarized as follows:

\underline {Comparison with interpretable algorithms}: Our algorithm demonstrates superior performance in terms of F1-score and NMI compared to IMM and SHA, while exhibiting comparable performance with respect to Purity.     As illustrated in Table~\ref{table6}, the decision tree generated by our method is slightly more compact than IMM yet marginally larger than SHA.     These results collectively indicate that our approach achieves enhanced clustering accuracy without compromising interpretability, outperforming existing interpretable clustering methods in regard to clustering quality.

\underline{Comparison with clustering ensemble algorithms}: Although our tree-based interpretable clustering ensemble algorithm does not achieve the best overall clustering performance, it exhibits competitive accuracy and shows particular advantages on specific data sets.    This observation reinforces the inherent benefits of clustering ensemble strategies, where our method maintains sufficient accuracy through tree-based ensemble model while preserving interpretability.

\begin{table*}[htbp]
\caption{The comparison of different clustering algorithms in terms of F1-score.}
\renewcommand{\arraystretch}{1}
\label{table4}
\centering
\begin{tabularx}{0.8\textwidth}{c|c|c|cc|ccccc}
\hline
Data set&ICE&KM&IMM&SHA&LWEA&LWGP&ECCMS&ECPCS-HC&ECPCS-MC \\\hline
Appendicitis  & 0.8033  & 0.7517  & 0.7474  & 0.8033  & 0.7655  & 0.7492  & 0.7655  & 0.7361  & 0.7233   \\ 
        Banana  & 0.5109  & 0.5126  & 0.5096  & 0.5098  & 0.5147  & 0.5156  & 0.5350  & 0.5181  & 0.5152   \\ 
        Bands  & 0.6712  & 0.5488  & 0.5386  & 0.5578  & 0.5702  & 0.6277  & 0.6019  & 0.6836  & 0.6836   \\ 
        Banknote  & 0.5966  & 0.5095  & 0.5410  & 0.5446  & 0.5459  & 0.5559  & 0.5817  & 0.5312  & 0.5258   \\ 
        Cervical Cancer & 0.6471  & 0.6896  & 0.6506  & 0.7127  & 0.6816  & 0.6328  & 0.6816  & 0.6956  & 0.6804   \\ 
        Glass   & 0.3820  & 0.3937  & 0.4111  & 0.3906  & 0.4090  & 0.4036  & 0.4104  & 0.4235  & 0.4067   \\ 
        Haberman  & 0.5487  & 0.5485  & 0.5687  & 0.5525  & 0.6083  & 0.5682  & 0.6263  & 0.6840  & 0.6231   \\ 
        Ionosphere  & 0.5964  & 0.6007  & 0.5936  & 0.5964  & 0.6128  & 0.6002  & 0.6119  & 0.5994  & 0.6024   \\ 
        Iris  & 0.8051  & 0.7408  & 0.7102  & 0.7319  & 0.7524  & 0.7463  & 0.7462  & 0.7471  & 0.7404   \\ 
        Maternal   & 0.5247  & 0.4416  & 0.4821  & 0.4338  & 0.4964  & 0.5093  & 0.4967  & 0.4632  & 0.5068   \\ 
        Movement  & 0.3285  & 0.3534  & 0.2876  & 0.3157  & 0.4031  & 0.4032  & 0.4031  & 0.4181  & 0.3967   \\ 
        Newthyroid  & 0.8336  & 0.8336  & 0.8336  & 0.8336  & 0.8403  & 0.8432  & 0.8414  & 0.8379  & 0.8640   \\ 
        NHANES  & 0.6894  & 0.7024  & 0.7135  & 0.7110  & 0.7091  & 0.6361  & 0.6842  & 0.7454  & 0.7396   \\ 
        Page-blocks  & 0.7543  & 0.5390  & 0.6561  & 0.7100  & 0.6704  & 0.5068  & 0.8616  & 0.7812  & 0.6389   \\ 
        Parkinsons  & 0.7166  & 0.6407  & 0.6449  & 0.6369  & 0.6657  & 0.6153  & 0.6787  & 0.6832  & 0.6291   \\ 
        Penbased  & 0.5283  & 0.5909  & 0.5582  & 0.5058  & 0.6775  & 0.7004  & 0.6534  & 0.6546  & 0.6634   \\ 
        Phoneme  & 0.6317  & 0.5985  & 0.6025  & 0.5732  & 0.6114  & 0.6234  & 0.6139  & 0.6150  & 0.6127   \\
         Pima  & 0.6652  & 0.5950  & 0.5996  & 0.5741  & 0.6062  & 0.6123  & 0.6059  & 0.6239  & 0.6275   \\ 
        Ring  & 0.6209  & 0.6555  & 0.5667  & 0.5675  & 0.6593  & 0.6811  & 0.6619  & 0.6750  & 0.6726   \\ 
        Satimage  & 0.5956  & 0.6166  & 0.5766  & 0.5660  & 0.6404  & 0.5516  & 0.5886  & 0.6430  & 0.5655   \\ 
        Seeds  & 0.7406  & 0.8482  & 0.7944  & 0.7678  & 0.7936  & 0.7841  & 0.7857  & 0.7973  & 0.7452   \\ 
        Segment  & 0.5102  & 0.5659  & 0.5949  & 0.5695  & 0.5177  & 0.5367  & 0.5033  & 0.5201  & 0.5374   \\ 
        Sonar  & 0.5662  & 0.5559  & 0.5619  & 0.5925  & 0.5715  & 0.5672  & 0.5740  & 0.5665  & 0.5626   \\ 
        Spambase  & 0.6814  & 0.6812  & 0.6717  & 0.6812  & 0.6818  & 0.6812  & 0.6816  & 0.6815  & 0.6812   \\ 
        Spectfheart  & 0.7354  & 0.6888  & 0.7248  & 0.7203  & 0.7037  & 0.6716  & 0.7111  & 0.6565  & 0.6786   \\ 
        Texture  & 0.5402  & 0.5524  & 0.5091  & 0.5254  & 0.6121  & 0.5630  & 0.6046  & 0.5924  & 0.5612   \\ 
        User knowledge   & 0.5114  & 0.3595  & 0.4312  & 0.4208  & 0.4052  & 0.4200  & 0.4337  & 0.3891  & 0.4039   \\ 
        Vehicle  & 0.3383  & 0.3213  & 0.3185  & 0.3370  & 0.3449  & 0.3359  & 0.3470  & 0.3556  & 0.3424   \\ 
        Wdbc  & 0.8013  & 0.8470  & 0.8198  & 0.8343  & 0.8353  & 0.8548  & 0.8247  & 0.8411  & 0.8390   \\ 
        Wine  & 0.8323  & 0.9296  & 0.8289  & 0.8324  & 0.9216  & 0.9360  & 0.9174  & 0.9361  & 0.9348   \\ 
        Winequality-red  & 0.3901  & 0.3299  & 0.3267  & 0.3261  & 0.3368  & 0.3305  & 0.3446  & 0.3876  & 0.3372   \\ 
        Wisconsin  & 0.8826  & 0.9239  & 0.8826  & 0.8826  & 0.9473  & 0.9470  & 0.9475  & 0.9468  & 0.9439   \\ 
        Yeast  & 0.3472  & 0.3067  & 0.3230  & 0.3086  & 0.3149  & 0.3144  & 0.3094  & 0.4048  & 0.3864   \\ \hline
        Avg & 0.6160  & 0.5992  & 0.5933  & 0.5947  & 0.6190  & 0.6068  & 0.6253  & 0.6313  & 0.6173   \\
         
\hline
\end{tabularx}
\end{table*}
\begin{figure*}[htbp]
  \centering

  \begin{minipage}{\textwidth}
    \centering
    \begin{tikzpicture}[
      level 1/.style={sibling distance=4.5cm, level distance=0.8cm},
      level 2/.style={sibling distance=2.5cm, level distance=0.8cm},
      every node/.style={draw, rounded corners, align=center, fill=gray!10, font=\scriptsize},
      edge from parent path={(\tikzparentnode.south) -- ++(0cm,-0.2cm) -| (\tikzchildnode.north)}
    ]
      \node {$f_4 \leq 0.47$}
        child {node {$f_3 \leq 0.5$}
          child {node[fill=red!30] {\#samples=121}}
          child {node[fill=red!30] {\#samples=87}}
        }
        child {node {$f_1 \leq 0.58$}
          child {node[fill=red!30] {\#samples=148}}
          child {node[fill=red!30] {\#samples=47}}
        };
    \end{tikzpicture}

    \vspace{0.3cm}
    (a) ICE
  \end{minipage}

  \vspace{0.5cm}

  \begin{minipage}{\textwidth}
    \centering
    \begin{tikzpicture}[
      level 1/.style={sibling distance=4cm, level distance=0.8cm},
      level 2/.style={sibling distance=2.5cm, level distance=0.8cm},
      every node/.style={draw, rounded corners, align=center, fill=gray!10, font=\scriptsize},
      edge from parent path={(\tikzparentnode.south) -- ++(0cm,-0.2cm) -| (\tikzchildnode.north)}
    ]
      \node {$f_1 \leq0.59$}
        child {node {$f_4\leq 0.48$}
          child {node {$f_2\leq 0.51$}
            child {node[fill=red!30] {\#samples=110}}
            child {node[fill=red!30] {\#samples=72}}
          }
          child {node[fill=red!30] {\#samples=148}}
        }
        child {node[fill=red!30] {\#samples=73}};
    \end{tikzpicture}

    \vspace{0.3cm}
    (b) IMM
  \end{minipage}

  \vspace{0.5cm}

  \begin{minipage}{\textwidth}
    \centering
    \begin{tikzpicture}[
      level 1/.style={sibling distance=4cm, level distance=0.8cm},
      level 2/.style={sibling distance=2.5cm, level distance=0.8cm},
      every node/.style={draw, rounded corners, align=center, fill=gray!10, font=\scriptsize},
      edge from parent path={(\tikzparentnode.south) -- ++(0cm,-0.2cm) -| (\tikzchildnode.north)}
    ]
      \node {$f_0 \leq0.55$}
        child {node {$f_3\leq 0.56$}
          child {node {$f_4\leq 0.45$}
            child {node[fill=red!30] {\#samples=106}}
            child {node[fill=red!30] {\#samples=117}}
          }
          child {node[fill=red!30] {\#samples=102}}
        }
        child {node[fill=red!30] {\#samples=78}};
    \end{tikzpicture}

    \vspace{0.3cm}
    (c) SHA
  \end{minipage}

  \caption{Decision trees generated by ICE, IMM and SHA.}
  \label{fig:decision-trees}
\end{figure*}

\begin{table*}[!t]
\caption{The comparison of different clustering algorithms in terms of NMI.}
\renewcommand{\arraystretch}{1}
\label{table5}
\centering
\adjustbox{max width=\textwidth}{
\begin{tabularx}{0.8\textwidth}{c|c|c|cc|ccccc}
\hline
Data set&ICE&KM&IMM&SHA&LWEA&LWGP&ECCMS&ECPCS-HC&ECPCS-MC \\\hline
Appendicitis                  & 0.2630 & 0.1944 & 0.2113 & 0.2630 & 0.1861 & 0.1909 & 0.1861 & 0.1781 & 0.1639 \\
Banana                        & 0.0122 & 0.0146 & 0.0100 & 0.0104 & 0.0025 & 0.0066 & 0.0099 & 0.0061 & 0.0113 \\
Bands                         & 0.0134 & 0.0037 & 0.0001 & 0.0001 & 0.0023 & 0.0097 & 0.0049 & 0.0148 & 0.0148 \\
Banknote    & 0.0400 & 0.0109 & 0.0509 & 0.0649 & 0.0317 & 0.0353 & 0.0549 & 0.0231 & 0.0180 \\
Cervical Cancer  & 0.2180 & 0.2400 & 0.1988 & 0.2671 & 0.2248 & 0.1594 & 0.2248 & 0.2467 & 0.2296 \\
Glass                         & 0.3178 & 0.3132 & 0.3286 & 0.3139 & 0.3542 & 0.3480 & 0.3561 & 0.3585 & 0.3499 \\
Haberman                      & 0.0007 & 0.0012 & 0.0071 & 0.0007 & 0.0210 & 0.0081 & 0.0257 & 0.0433 & 0.0231 \\
Ionosphere                    & 0.1059 & 0.1250 & 0.0960 & 0.1059 & 0.1446 & 0.1320 & 0.1428 & 0.1309 & 0.1327 \\
Iris                          & 0.7735 & 0.6553 & 0.6354 & 0.6461 & 0.6832 & 0.6667 & 0.6853 & 0.6789 & 0.6603 \\
Maternal  & 0.2143 & 0.1543 & 0.1940 & 0.1521 & 0.1797 & 0.1882 & 0.1797 & 0.1453 & 0.1980 \\
Movement                      & 0.5623 & 0.5886 & 0.4982 & 0.5436 & 0.6387 & 0.6318 & 0.6387 & 0.6540 & 0.6269 \\
Newthyroid                    & 0.5590 & 0.5590 & 0.5590 & 0.5590 & 0.5787 & 0.5855 & 0.5822 & 0.5712 & 0.6308 \\
NHANES     & 0.0026 & 0.0044 & 0.0018 & 0.0028 & 0.0018 & 0.0009 & 0.0022 & 0.0033 & 0.0023 \\
Page-blocks                   & 0.2067 & 0.0993 & 0.0706 & 0.1024 & 0.1960 & 0.1532 & 0.2756 & 0.2340 & 0.1268 \\
Parkinsons                    & 0.0457 & 0.1024 & 0.0942 & 0.1051 & 0.1359 & 0.2514 & 0.1030 & 0.0800 & 0.1461 \\
Penbased                      & 0.6061 & 0.6884 & 0.6315 & 0.6001 & 0.7509 & 0.7571 & 0.7434 & 0.7430 & 0.7497 \\
Phoneme                       & 0.1023 & 0.1761 & 0.1272 & 0.1205 & 0.1816 & 0.1237 & 0.1742 & 0.1812 & 0.1754 \\
Pima                          & 0.0109 & 0.0617 & 0.0509 & 0.0449 & 0.0257 & 0.0563 & 0.0294 & 0.0388 & 0.0519 \\
Ring                          & 0.0867 & 0.2542 & 0.0462 & 0.0475 & 0.2630 & 0.2986 & 0.2685 & 0.2810 & 0.2854 \\
Satimage                      & 0.5682 & 0.6130 & 0.5516 & 0.5575 & 0.6221 & 0.5948 & 0.5932 & 0.6311 & 0.5996 \\
Seeds                         & 0.6503 & 0.7279 & 0.6597 & 0.6226 & 0.6703 & 0.6622 & 0.6637 & 0.6733 & 0.6292 \\
Segment                       & 0.5622 & 0.6008 & 0.6357 & 0.6114 & 0.5605 & 0.5711 & 0.5536 & 0.5857 & 0.5895 \\
Sonar                         & 0.0381 & 0.0085 & 0.0138 & 0.0504 & 0.0195 & 0.0146 & 0.0229 & 0.0145 & 0.0098 \\
Spambase                      & 0.0099 & 0.0104 & 0.0615 & 0.0104 & 0.0091 & 0.0104 & 0.0097 & 0.0098 & 0.0104 \\
Spectfheart                   & 0.0466 & 0.0731 & 0.0280 & 0.0303 & 0.0646 & 0.0836 & 0.0604 & 0.0945 & 0.0791 \\
Texture                       & 0.6432 & 0.6365 & 0.6114 & 0.6156 & 0.6973 & 0.6787 & 0.7135 & 0.6850 & 0.6804 \\
User knowledge       & 0.4576 & 0.1928 & 0.2990 & 0.3049 & 0.2365 & 0.3147 & 0.2942 & 0.2181 & 0.2457 \\
Vehicle                       & 0.1146 & 0.1050 & 0.1086 & 0.1118 & 0.1216 & 0.1316 & 0.1241 & 0.1360 & 0.1141 \\
Wdbc                          & 0.4720 & 0.5480 & 0.4880 & 0.5186 & 0.5335 & 0.5652 & 0.5118 & 0.5361 & 0.5282 \\
Wine                          & 0.7095 & 0.8730 & 0.7006 & 0.7052 & 0.8601 & 0.8782 & 0.8546 & 0.8818 & 0.8747 \\
Winequality-red               & 0.1046 & 0.0890 & 0.0886 & 0.0743 & 0.0969 & 0.0964 & 0.0941 & 0.0851 & 0.0835 \\
Wisconsin                     & 0.6220 & 0.7283 & 0.6220 & 0.6220 & 0.7985 & 0.7976 & 0.7993 & 0.7971 & 0.7872 \\
Yeast                         & 0.2717 & 0.2948 & 0.2736 & 0.2806 & 0.2996 & 0.2898 & 0.2990 & 0.3250 & 0.3145 \\ \hline
Avg                           & 0.2852 & 0.2954 & 0.2713 & 0.2747 & 0.3089 & 0.3119 & 0.3116 & 0.3117 & 0.3074 \\
         
\hline
\end{tabularx}
}
\end{table*}

\begin{figure*}[!t]
  \centering

  \subfloat[Purity]{%
    \includegraphics[width=0.4\textwidth]{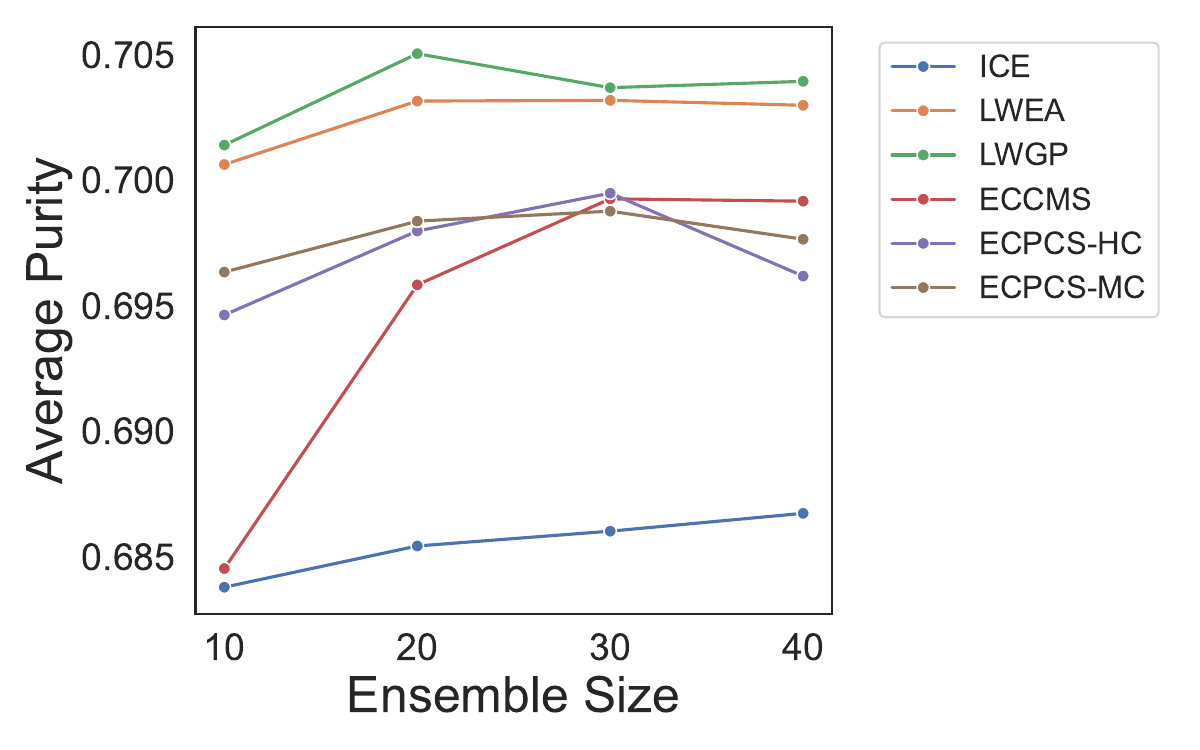}
    \label{fig:purity}
  }
  \hspace{0.02\textwidth}
  \subfloat[F1-score]{%
    \includegraphics[width=0.4\textwidth]{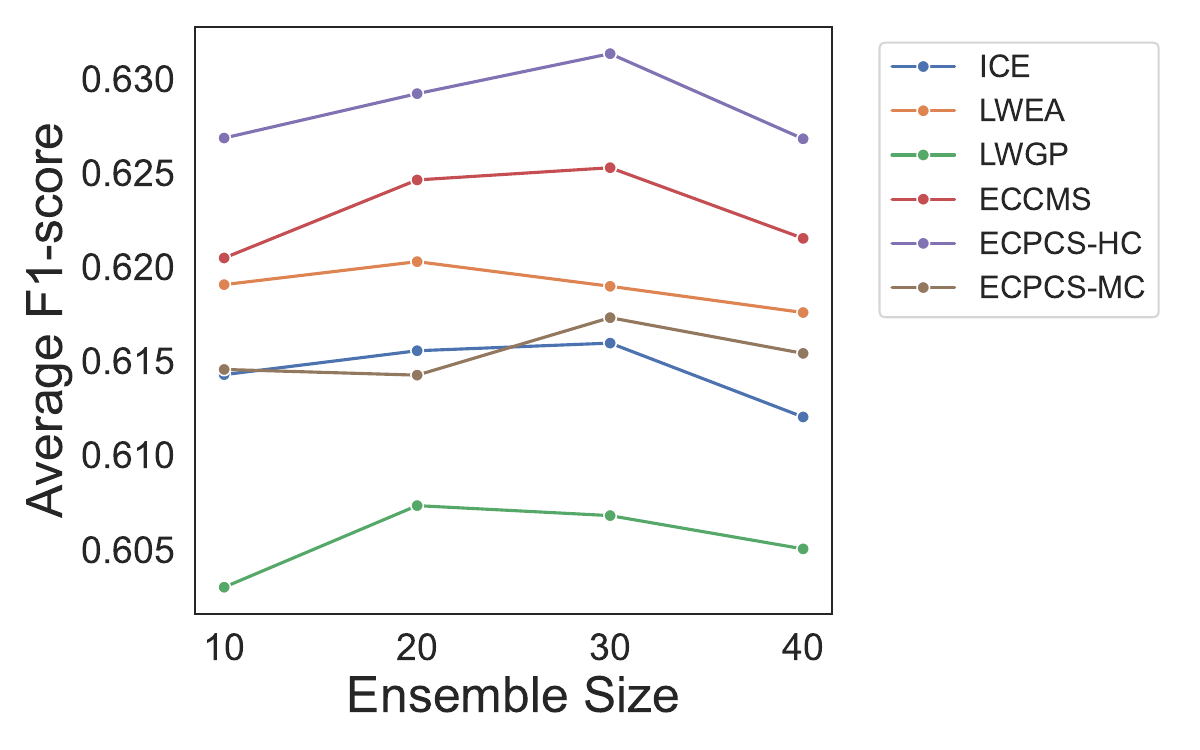}
    \label{fig:fscore}
  }
  \hspace{0.02\textwidth}
  \subfloat[NMI]{%
    \includegraphics[width=0.4\textwidth]{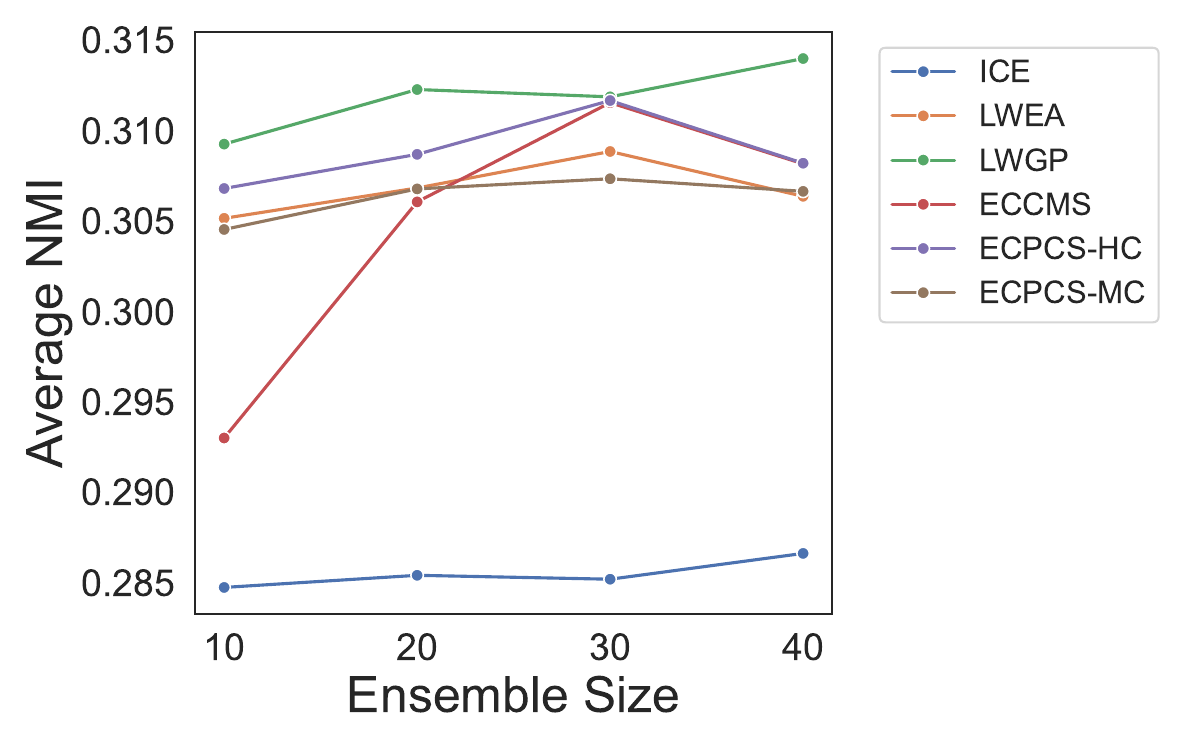}
    \label{fig:nmi}
  }

  \caption{Comparison of clustering performance metrics across different ensemble sizes.}
  \label{fig:ensemble-metrics}
\end{figure*}

\subsection{The interpretability comparison}
Since our algorithm is based on decision trees, we use the maximum depth of the tree (maxDepth) and the average depth of the leaf nodes (avgDepth) to evaluate the interpretability of each algorithm, as shown in Table~\ref{table6}. Although the number of leaf nodes is also an important metric for evaluating interpretability, our algorithm, as well as IMM and SHA, prespecify the number of leaf nodes to be the number of ground-truth clusters. Therefore, the number of leaf nodes is not used for the purpose of interpretability comparison.

The experimental results demonstrate that the ICE algorithm outperforms IMM in terms of both maximum tree depth and average depth, while slightly lagging behind SHA, indicating its ability to balance interpretability and clustering quality effectively.

To illustrate the structural differences among interpretable clustering algorithms more specifically, we visualize the decision trees generated by ICE, IMM, and SHA on the User knowledge dataset, as shown in Fig.~\ref{fig:decision-trees}. 

IMM and SHA exhibit similar tree depths and branching patterns, they differ significantly in their splitting conditions. For instance, IMM uses features $f_1,f_4,f_2,$ while SHA relies on $f_0,f_3,f_4$, indicating that each method prioritizes different aspects of the data. This variation arises from their distinct splitting criteria and tree-growth strategies. 

In contrast, ICE not only adopts different splitting features but also demonstrates a more balanced tree structure. This shows how ICE encourages more diverse and meaningful partitioning, potentially leading to enhanced interpretability.

\begin{table}[!t]
\caption{Comparison of interpretable clustering algorithms in terms of maxDepth and avgDepth.}
\renewcommand{\arraystretch}{1}
\label{table6}
\centering
\adjustbox{max width=\textwidth}{
\begin{tabular}{lccc@{\hskip 1em}ccc}
\toprule
\multirow{2}{*}{\textbf{Data Set}} & \multicolumn{3}{c}{\textbf{avgDepth}} & \multicolumn{3}{c}{\textbf{maxDepth}} \\
\cmidrule(lr){2-4} \cmidrule(lr){5-7}
& ICE & IMM & SHA & ICE & IMM & SHA \\\midrule
Appendicitis         & 1.00 & 1.00 & 1.00 & 1.00 & 1.00 & 1.00 \\
Banana               & 1.00 & 1.00 & 1.00 & 1.00 & 1.00 & 1.00 \\
Bands                & 1.00 & 1.00 & 1.00 & 1.00 & 1.00 & 1.00 \\
Banknote             & 1.00 & 1.00 & 1.00 & 1.00 & 1.00 & 1.00 \\
Cervical Cancer      & 1.00 & 1.00 & 1.00 & 1.00 & 1.00 & 1.00 \\
Glass                & 3.33 & 3.25 & 3.33 & 5.00 & 4.80 & 5.00 \\
Haberman             & 1.00 & 1.00 & 1.00 & 1.00 & 1.00 & 1.00 \\
Ionosphere           & 1.00 & 1.00 & 1.00 & 1.00 & 1.00 & 1.00 \\
Iris                 & 1.67 & 1.67 & 1.67 & 2.00 & 2.00 & 2.00 \\
Maternal             & 1.67 & 1.67 & 1.67 & 2.00 & 2.00 & 2.00 \\
Movement             & 4.81 & 7.19 & 4.23 & 7.20 & 11.30 & 5.60 \\
Newthyroid           & 1.67 & 1.67 & 1.67 & 2.00 & 2.00 & 2.00 \\
NHANES               & 1.00 & 1.00 & 1.00 & 1.00 & 1.00 & 1.00 \\
Page-blocks          & 2.80 & 2.80 & 2.48 & 4.00 & 4.00 & 3.10 \\
Parkinsons           & 1.00 & 1.00 & 1.00 & 1.00 & 1.00 & 1.00 \\
Penbased             & 3.82 & 4.01 & 3.59 & 5.20 & 6.40 & 4.90 \\
Phoneme              & 1.00 & 1.00 & 1.00 & 1.00 & 1.00 & 1.00 \\
Pima                 & 1.00 & 1.00 & 1.00 & 1.00 & 1.00 & 1.00 \\
Ring                 & 1.00 & 1.00 & 1.00 & 1.00 & 1.00 & 1.00 \\
Satimage             & 3.33 & 3.23 & 3.33 & 5.00 & 4.80 & 5.00 \\
Seeds                & 1.67 & 1.67 & 1.67 & 2.00 & 2.00 & 2.00 \\
Segment              & 3.86 & 3.86 & 3.86 & 6.00 & 6.00 & 6.00 \\
Sonar                & 1.00 & 1.00 & 1.00 & 1.00 & 1.00 & 1.00 \\
Spambase             & 1.00 & 1.00 & 1.00 & 1.00 & 1.00 & 1.00 \\
Spectfheart          & 1.00 & 1.00 & 1.00 & 1.00 & 1.00 & 1.00 \\
Texture              & 5.25 & 5.72 & 3.99 & 8.70 & 9.70 & 6.00 \\
User knowledge       & 2.00 & 2.25 & 2.25 & 2.00 & 3.00 & 3.00 \\
Vehicle              & 2.25 & 2.25 & 2.25 & 3.00 & 3.00 & 3.00 \\
Wdbc                 & 1.00 & 1.00 & 1.00 & 1.00 & 1.00 & 1.00 \\
Wine                 & 1.67 & 1.67 & 1.67 & 2.00 & 2.00 & 2.00 \\
Winequality-red      & 3.33 & 3.30 & 3.33 & 5.00 & 4.80 & 5.00 \\
Wisconsin            & 1.00 & 1.00 & 1.00 & 1.00 & 1.00 & 1.00 \\
Yeast                & 5.09 & 5.33 & 5.36 & 8.50 & 8.90 & 8.90 \\\midrule
Avg                  & 1.98 & 2.08 & 1.92 & 2.62 & 2.84 & 2.50 \\
\bottomrule
\end{tabular}
}
\end{table}

\subsection{The influence of ensemble size}

In this section, we evaluate the impact of the ensemble size on the performance of the ICE algorithm and other representative clustering ensemble methods. The ensemble size is set to 10, 20, 30, 40 and each method is run 10 times on 33 data sets to obtain its average performance.

Fig.~\ref{fig:ensemble-metrics} presents the average clustering performance across all data sets in terms of three evaluation metrics. We observe that as the ensemble size increases, the clustering performance generally improves, which aligns with our expectations—larger ensembles typically capture more diverse information, leading to better overall quality.

\subsection{Execution Time}
We evaluated the average execution time over 10 runs for the proposed ICE algorithm and baselines across all 33 datasets. The experimental results are shown in Fig.~\ref{figb}. Overall, ICE incurs a higher time cost, primarily because it exhaustively considers all possible candidate splits when determining the best partition for a candidate node. Although the algorithm has been optimized through incremental updates, it remains computationally intensive. Despite of its high computational cost, this trade-off is acceptable given that our method outperforms other tree-based interpretable algorithms in terms of clustering performance.
\section{Conclusions}
In this paper, we propose ICE, a novel decision tree based clustering algorithm designed to bring interpretability to clustering ensemble. By leveraging statistical association test at each node, ICE chooses the candidate split that can achieve the highest association with base partitions to grow the decision tree. Experimental results on 33 datasets demonstrate that ICE achieves comparable clustering performance  to SOTA clustering ensemble models while maintaining clear interpretability, and outperforms existing interpretable clustering methods in terms of clustering accuracy.

However, the ICE algorithm still has several limitations that warrant further investigation. First, it incurs high computational cost due to the exhaustive evaluation of all candidate splits. Second, the current strategy of selecting the split with the smallest $p$-value may not always yield the most optimal partition.

For future research, pruning strategies could be introduced to reduce the computational burden by focusing only on branches that are more likely to yield optimal splits. Moreover, relying solely on $p$-value as the statistical criterion for split selection has inherent limitations. Future work may explore alternative statistical methods  to enhance the robustness of ICE.

\section*{Acknowledgments}

This work has been supported by the Natural Science Foundation of China under Grant No. 62472064.
\bibliographystyle{IEEEtran}
\bibliography{cite}

\begin{thebibliography}{10}
\providecommand{\url}[1]{#1}
\csname url@samestyle\endcsname
\providecommand{\newblock}{\relax}
\providecommand{\bibinfo}[2]{#2}
\providecommand{\BIBentrySTDinterwordspacing}{\spaceskip=0pt\relax}
\providecommand{\BIBentryALTinterwordstretchfactor}{4}
\providecommand{\BIBentryALTinterwordspacing}{\spaceskip=\fontdimen2\font plus
\BIBentryALTinterwordstretchfactor\fontdimen3\font minus \fontdimen4\font\relax}
\providecommand{\BIBforeignlanguage}[2]{{%
\expandafter\ifx\csname l@#1\endcsname\relax
\typeout{** WARNING: IEEEtran.bst: No hyphenation pattern has been}%
\typeout{** loaded for the language `#1'. Using the pattern for}%
\typeout{** the default language instead.}%
\else
\language=\csname l@#1\endcsname
\fi
#2}}
\providecommand{\BIBdecl}{\relax}
\BIBdecl

\bibitem{oyewole2023data}
G.~J. Oyewole and G.~A. Thopil, ``Data clustering: application and trends,'' \emph{Artificial intelligence review}, vol.~56, no.~7, pp. 6439--6475, 2023.

\bibitem{Mittal2022}
H.~Mittal, A.~C. Pandey, M.~Saraswat, S.~Kumar, R.~Pal, and G.~Modwel, ``A comprehensive survey of image segmentation: clustering methods, performance parameters, and benchmark datasets,'' \emph{Multimedia Tools and Applications}, vol.~81, no.~24, pp. 35\,001--35\,026, Oct 2022.

\bibitem{Socialnet}
W.-L. Shiau, Y.~K. Dwivedi, and H.~S. Yang, ``Co-citation and cluster analyses of extant literature on social networks,'' \emph{International Journal of Information Management}, vol.~37, no.~5, pp. 390--399, 2017.

\bibitem{gene}
D.~Jiang, C.~Tang, and A.~Zhang, ``Cluster analysis for gene expression data: a survey,'' \emph{IEEE Transactions on Knowledge and Data Engineering}, vol.~16, no.~11, pp. 1370--1386, 2004.

\bibitem{GOLALIPOUR2021104388}
K.~Golalipour, E.~Akbari, S.~S. Hamidi, M.~Lee, and R.~Enayatifar, ``From clustering to clustering ensemble selection: A review,'' \emph{Engineering Applications of Artificial Intelligence}, vol. 104, p. 104388, 2021.

\bibitem{Hu2024InterpretableCA}
L.~Hu, M.~Jiang, J.~Dong, X.~Liu, and Z.~He, ``Interpretable clustering: A survey,'' \emph{ArXiv}, vol. abs/2409.00743, 2024.

\bibitem{SHA}
E.~Laber, L.~Murtinho, and F.~Oliveira, ``Shallow decision trees for explainable k-means clustering,'' \emph{Pattern Recognition}, vol. 137, p. 109239, 2023.

\bibitem{bertsimas2021interpretable}
D.~Bertsimas, A.~Orfanoudaki, and H.~Wiberg, ``Interpretable clustering: an optimization approach,'' \emph{Machine Learning}, vol. 110, no.~1, pp. 89--138, 2021.

\bibitem{2011survey}
S.~Vega-Pons and J.~Ruiz-Shulcloper, ``A survey of clustering ensemble algorithms,'' \emph{International Journal of Pattern Recognition and Artificial Intelligence}, vol.~25, no.~03, pp. 337--372, 2011.

\bibitem{BOONGOEN2018}
T.~Boongoen and N.~Iam-On, ``Cluster ensembles: A survey of approaches with recent extensions and applications,'' \emph{Computer Science Review}, vol.~28, pp. 1--25, 2018.

\bibitem{Strehl2002}
A.~Strehl and J.~Ghosh, ``Cluster ensembles --- {A} knowledge reuse framework for combining multiple partitions,'' \emph{J. Mach. Learn. Res.}, vol.~3, pp. 583--617, 2002.

\bibitem{fern2004solving}
X.~Z. Fern and C.~E. Brodley, ``Solving cluster ensemble problems by bipartite graph partitioning,'' in \emph{Proceedings of the twenty-first international conference on Machine learning}, 2004, p.~36.

\bibitem{fred2005combining}
A.~L. Fred and A.~K. Jain, ``Combining multiple clusterings using evidence accumulation,'' \emph{IEEE transactions on pattern analysis and machine intelligence}, vol.~27, no.~6, pp. 835--850, 2005.

\bibitem{abdala2010ensemble}
D.~D. Abdala, P.~Wattuya, and X.~Jiang, ``Ensemble clustering via random walker consensus strategy,'' in \emph{2010 20th International Conference on Pattern Recognition}.\hskip 1em plus 0.5em minus 0.4em\relax IEEE, 2010, pp. 1433--1436.

\bibitem{wang2009clustering}
X.~Wang, C.~Yang, and J.~Zhou, ``Clustering aggregation by probability accumulation,'' \emph{Pattern Recognition}, vol.~42, no.~5, pp. 668--675, 2009.

\bibitem{zhou2023active}
P.~Zhou, B.~Sun, X.~Liu, L.~Du, and X.~Li, ``Active clustering ensemble with self-paced learning,'' \emph{IEEE Transactions on Neural Networks and Learning Systems}, 2023.

\bibitem{li2024adaptive}
T.~Li, X.~Shu, J.~Wu, Q.~Zheng, X.~Lv, and J.~Xu, ``Adaptive weighted ensemble clustering via kernel learning and local information preservation,'' \emph{Knowledge-Based Systems}, vol. 294, p. 111793, 2024.

\bibitem{khedairia2022multiple}
S.~Khedairia and M.~T. Khadir, ``A multiple clustering combination approach based on iterative voting process,'' \emph{Journal of King Saud University-Computer and Information Sciences}, vol.~34, no.~1, pp. 1370--1380, 2022.

\bibitem{ayad2010voting}
H.~G. Ayad and M.~S. Kamel, ``On voting-based consensus of cluster ensembles,'' \emph{Pattern Recognition}, vol.~43, no.~5, pp. 1943--1953, 2010.

\bibitem{saeed2012voting}
F.~Saeed, N.~Salim, and A.~Abdo, ``Voting-based consensus clustering for combining multiple clusterings of chemical structures,'' \emph{Journal of cheminformatics}, vol.~4, pp. 1--8, 2012.

\bibitem{carrizosa2023clustering}
E.~Carrizosa, K.~Kurishchenko, A.~Mar{\'\i}n, and D.~R. Morales, ``On clustering and interpreting with rules by means of mathematical optimization,'' \emph{Computers \& Operations Research}, vol. 154, p. 106180, 2023.

\bibitem{fraiman2013interpretable}
R.~Fraiman, B.~Ghattas, and M.~Svarc, ``Interpretable clustering using unsupervised binary trees,'' \emph{Advances in Data Analysis and Classification}, vol.~7, pp. 125--145, 2013.

\bibitem{hwang2023xclusters}
H.~Hwang and S.~E. Whang, ``Xclusters: explainability-first clustering,'' in \emph{Proceedings of the AAAI Conference on Artificial Intelligence}, vol.~37, no.~7, 2023, pp. 7962--7970.

\bibitem{carrizosa2022interpreting}
E.~Carrizosa, K.~Kurishchenko, A.~Mar{\'\i}n, and D.~R. Morales, ``Interpreting clusters via prototype optimization,'' \emph{Omega}, vol. 107, p. 102543, 2022.

\bibitem{lawless2023cluster}
C.~Lawless and O.~Gunluk, ``Cluster explanation via polyhedral descriptions,'' in \emph{International conference on machine learning}.\hskip 1em plus 0.5em minus 0.4em\relax PMLR, 2023, pp. 18\,652--18\,666.

\bibitem{davidson2018cluster}
I.~Davidson, A.~Gourru, and S.~Ravi, ``The cluster description problem-complexity results, formulations and approximations,'' \emph{Advances in Neural Information Processing Systems}, vol.~31, 2018.

\bibitem{bonifati2022time2feat}
A.~Bonifati, F.~Del~Buono, F.~Guerra, and D.~Tiano, ``Time2feat: Learning interpretable representations for multivariate time series clustering,'' \emph{Proceedings of the VLDB Endowment (PVLDB)}, vol.~16, no.~2, pp. 193--201, 2022.

\bibitem{svirsky2023interpretable}
J.~Svirsky and O.~Lindenbaum, ``Interpretable deep clustering for tabular data,'' \emph{arXiv preprint arXiv:2306.04785}, 2023.

\bibitem{effenberger2021interpretable}
T.~Effenberger and R.~Pel{\'a}nek, ``Interpretable clustering of students’ solutions in introductory programming,'' in \emph{International Conference on Artificial Intelligence in Education}.\hskip 1em plus 0.5em minus 0.4em\relax Springer, 2021, pp. 101--112.

\bibitem{saisubramanian2020balancing}
S.~Saisubramanian, S.~Galhotra, and S.~Zilberstein, ``Balancing the tradeoff between clustering value and interpretability,'' in \emph{Proceedings of the AAAI/ACM Conference on AI, Ethics, and Society}, 2020, pp. 351--357.

\bibitem{pmlr-v119-moshkovitz20a}
M.~Moshkovitz, S.~Dasgupta, C.~Rashtchian, and N.~Frost, ``Explainable k-means and k-medians clustering,'' in \emph{Proceedings of the 37th International Conference on Machine Learning}, ser. Proceedings of Machine Learning Research, H.~D. III and A.~Singh, Eds., vol. 119.\hskip 1em plus 0.5em minus 0.4em\relax PMLR, 13--18 Jul 2020, pp. 7055--7065.

\bibitem{adolfsson2019cluster}
A.~Adolfsson, M.~Ackerman, and N.~C. Brownstein, ``To cluster, or not to cluster: An analysis of clusterability methods,'' \emph{Pattern Recognition}, vol.~88, pp. 13--26, 2019.

\bibitem{maitra2012bootstrapping}
R.~Maitra, V.~Melnykov, and S.~N. Lahiri, ``Bootstrapping for significance of compact clusters in multidimensional datasets,'' \emph{Journal of the American Statistical Association}, vol. 107, no. 497, pp. 378--392, 2012.

\bibitem{helgeson2021nonparametric}
E.~S. Helgeson, D.~M. Vock, and E.~Bair, ``Nonparametric cluster significance testing with reference to a unimodal null distribution,'' \emph{Biometrics}, vol.~77, no.~4, pp. 1215--1226, 2021.

\bibitem{xie2021significant}
Y.~Xie, X.~Jia, S.~Shekhar, H.~Bao, and X.~Zhou, ``Significant dbscan+: Statistically robust density-based clustering,'' \emph{ACM Transactions on Intelligent Systems and Technology (TIST)}, vol.~12, no.~5, pp. 1--26, 2021.

\bibitem{hu2025interpretable}
L.~Hu, M.~Jiang, J.~Dong, X.~Liu, and Z.~He, ``Interpretable categorical data clustering via hypothesis testing,'' \emph{Pattern Recognition}, p. 111364, 2025.

\bibitem{hu2025significance}
L.~Hu, M.~Jiang, X.~Liu, and Z.~He, ``Significance-based decision tree for interpretable categorical data clustering,'' \emph{Information Sciences}, vol. 690, p. 121588, 2025.

\bibitem{he2025significance}
Z.~He, L.~Hu, J.~He, J.~Dong, M.~Jiang, and X.~Liu, ``Significance-based interpretable sequence clustering,'' \emph{Information Sciences}, p. 121972, 2025.

\bibitem{agresti2013categorical}
A.~Agresti, \emph{Categorical data analysis}.\hskip 1em plus 0.5em minus 0.4em\relax John Wiley \& Sons, 2013.

\bibitem{huang2017locally}
D.~Huang, C.-D. Wang, and J.-H. Lai, ``Locally weighted ensemble clustering,'' \emph{IEEE transactions on cybernetics}, vol.~48, no.~5, pp. 1460--1473, 2017.

\bibitem{jia2023ensemble}
Y.~Jia, S.~Tao, R.~Wang, and Y.~Wang, ``Ensemble clustering via co-association matrix self-enhancement,'' \emph{IEEE Transactions on Neural Networks and Learning Systems}, 2023.

\bibitem{huang2018enhanced}
D.~Huang, C.-D. Wang, H.~Peng, J.~Lai, and C.-K. Kwoh, ``Enhanced ensemble clustering via fast propagation of cluster-wise similarities,'' \emph{IEEE Transactions on Systems, Man, and Cybernetics: Systems}, vol.~51, no.~1, pp. 508--520, 2018.

\bibitem{dua2017uci}
D.~Dua, C.~Graff \emph{et~al.}, ``Uci machine learning repository,'' 2017.

\bibitem{derrac2015keel}
J.~Derrac, S.~Garcia, L.~Sanchez, and F.~Herrera, ``Keel data-mining software tool: Data set repository, integration of algorithms and experimental analysis framework,'' \emph{J. Mult. Valued Logic Soft Comput}, vol.~17, pp. 255--287, 2015.

\end{thebibliography}

\begin{IEEEbiography}[{\includegraphics
[width=1in,height=1.25in,clip,
keepaspectratio]{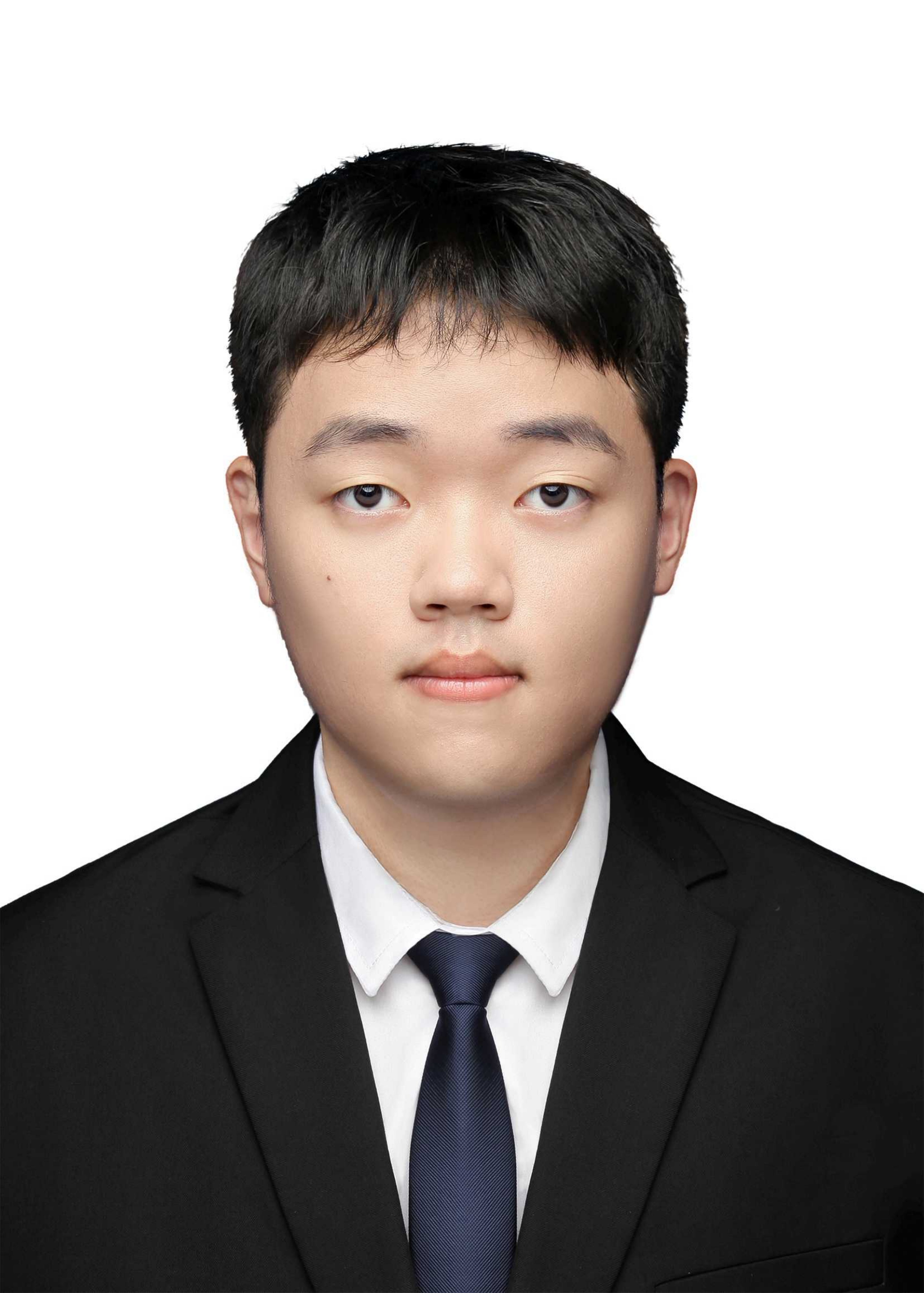}}]
{Hang Lv} 
received the BS degree in computer science from North China Electric Power University, Baoding, China, in 2024. He is currently working toward the MS degree with the School of Software, Dalian University of Technology, China. His current research interests include data mining and its applications.
\end{IEEEbiography}

\vspace{30pt}

\begin{IEEEbiography}[{\includegraphics[width=1in,height=1.25in,clip,keepaspectratio]{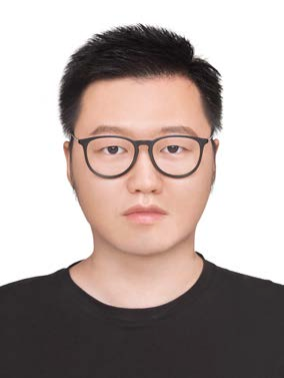}}]{Lianyu Hu}
received the MS degree in computer science from Ningbo University, China, in 2019. He is currently working toward the PhD degree in the School of Software at Dalian University of Technology. His current research interests include machine learning, cluster analysis and data mining.
\end{IEEEbiography}
\vspace{20mm}
\begin{IEEEbiography}[{\includegraphics[width=1in,height=1.25in,clip,keepaspectratio]{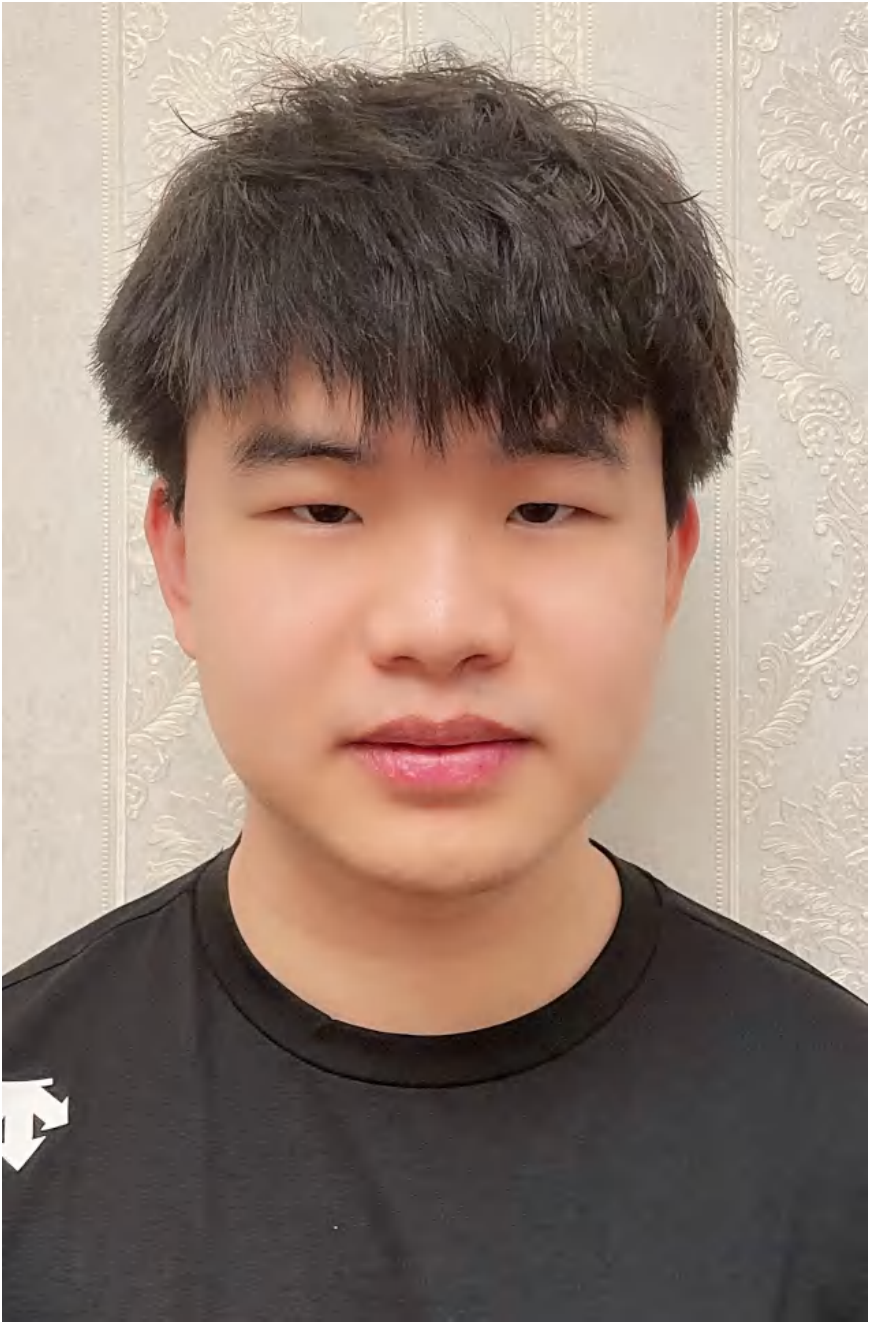}}]{Mudi Jiang}
received the MS degree in software engineering from Dalian University of Technology, China, in 2023. He is currently working toward the PhD degree in the School of Software at the same university. His current research interests include data mining and its applications.
\end{IEEEbiography}

\vspace{20mm}

\begin{IEEEbiography}[{\includegraphics[width=1in,height=1.25in,clip,keepaspectratio]{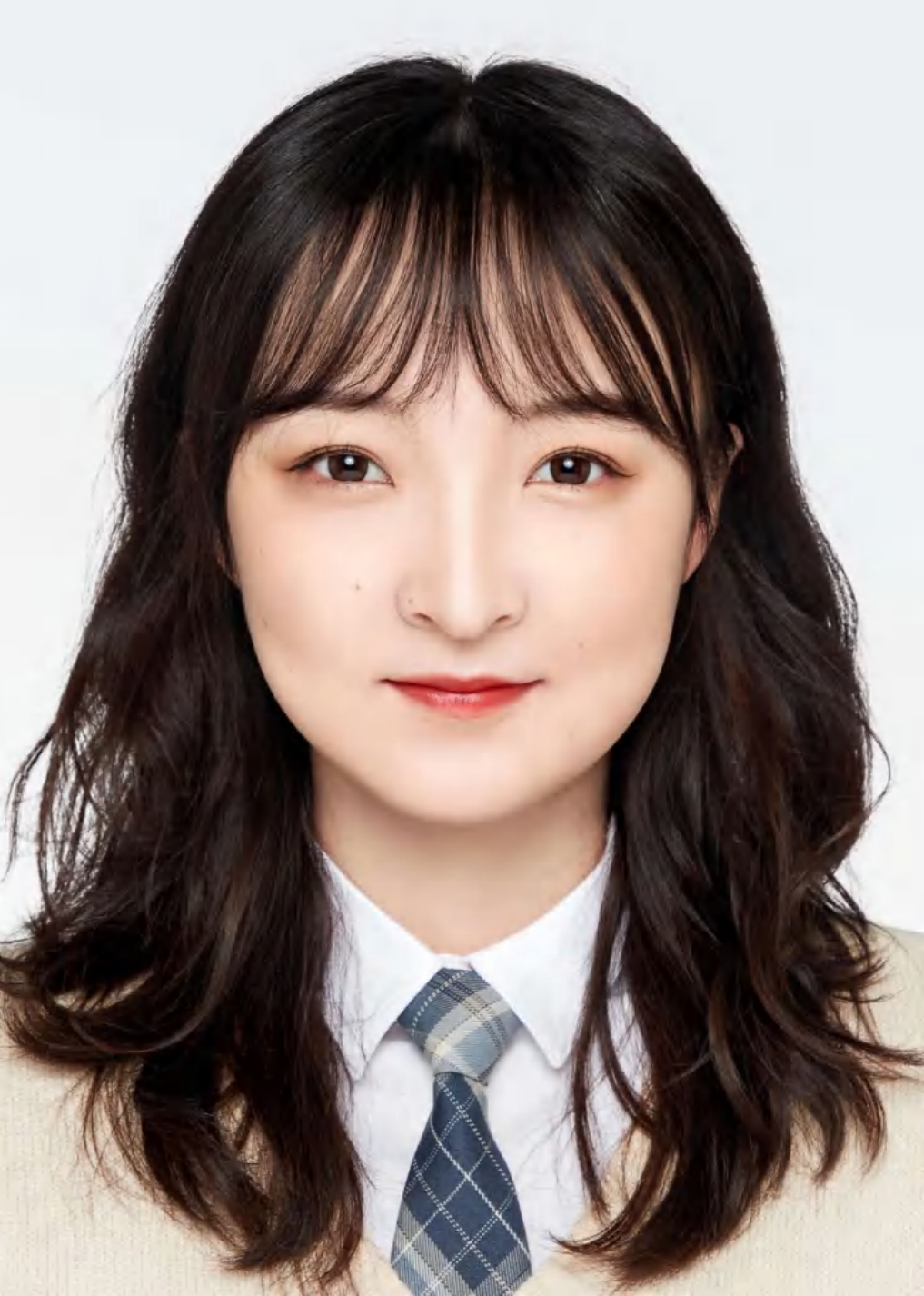}}]{Xinying Liu}
received the MS degree in Applied Statistics from China University of Geosciences, China, in 2023. She is currently working toward the PhD degree in the School of Software at Dalian University of Technology. Her current research interests include machine learning and data mining.
\end{IEEEbiography}
\vspace{-125mm}

\begin{IEEEbiography}[{\includegraphics[width=1in,height=1.25in,clip,keepaspectratio]{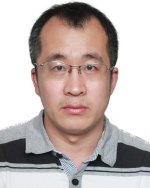}}]{Zengyou He}
received the BS, MS, and PhD degrees in computer science from Harbin Institute of Technology, China, in 2000, 2002, and 2006, respectively. He was a research associate in the Department of Electronic and Computer Engineering, Hong Kong University of Science and Technology from February 2007 to February 2010. He is currently a professor in the School of software, Dalian University of Technology. His research interests include data mining and bioinformatics.
\end{IEEEbiography}

\end{document}